\documentclass[journal,twoside,web]{ieeecolor}
\usepackage{tmi}
\usepackage{cite}
\usepackage{amsmath,amssymb,amsfonts}
\usepackage{algorithmic}
\usepackage{graphicx}
\usepackage{textcomp}
\usepackage{multirow, multicol, booktabs}
\usepackage[table, dvipsnames]{xcolor}

\def\BibTeX{{\rm B\kern-.05em{\sc i\kern-.025em b}\kern-.08em
    T\kern-.1667em\lower.7ex\hbox{E}\kern-.125emX}}
\begin{document}
\title{Federated Self-Supervised Learning for One-Shot Cross-Modal and Cross-Imaging Technique Segmentation}

\author{Siladittya Manna$^*$, Suresh Das$^*$, 
Sayantari Ghosh and Saumik Bhattacharya
\thanks{Siladittya Manna is with the Hong Kong Baptist University, Hong Kong when working on this research work. Email: smanna@hkbu.edu.hk}
\thanks{Suresh Das is  affiliated with Narayana Superspeciality Hospital, Howrah, India}
% \thanks{Prasun Sanki is affiliated with the Department of Electrical Engineering, Netaji Subhash Engineering College, Kolkata, India}
\thanks{Sayantari Ghosh is an Assistant Professor at the Department of Physics, National Institute of Technology Durgapur, India}
\thanks{Saumik Bhattacharya is an Assistant Professor at the Department of Electronics and Electrical Engineering, Indian Institute of Technology, Kharagpur. Email: saumik@ece.iitkgp.ac.in}
\thanks{$^*$ These authors have contributed equally.}}

\maketitle

\begin{abstract}
Decentralized federated learning enables learning of data representations from multiple sources without compromising the privacy of the clients. In applications like medical image segmentation, where obtaining a large annotated dataset from a single source is a distressing problem, federated self-supervised learning can provide some solace. In this work, we push the limits further by exploring a federated self-supervised one-shot segmentation task representing a more data-scarce scenario. We adopt a pre-existing self-supervised few-shot segmentation framework CoWPro and adapt it to the federated learning scenario. To the best of our knowledge, this work is the first to attempt a self-supervised few-shot segmentation task in the federated learning domain. Moreover, we consider the clients to be constituted of data from different modalities and imaging techniques like MR or CT, which makes the problem even harder. Additionally, we reinforce and improve the baseline CoWPro method using a fused dice loss which shows considerable improvement in performance over the baseline CoWPro. Finally, we evaluate this novel framework on a completely unseen held-out part of the local client dataset. We observe that the proposed framework can achieve performance at par or better than the FedAvg version of the CoWPro framework on the held-out validation dataset.
\end{abstract}


\section{Introduction}
\label{sec:intro}

%%%%%%%%% INTRODUCE FEDERATED LEARNING %%%%%%%%
\IEEEPARstart{T}{he} paradigm of federated learning (FL) provides the tools for representations from multiple entities called clients in the FL terminology, without hampering the privacy of those entities \cite{yang2019FedMLca, Wen2022fedlsurvey}. First introduced in 2016 by Google, the primary idea of federated learning was aimed at learning model representations locally on mobile devices without sharing user data and preserving user privacy \cite{mcmahan17afedavg}. This first work FedAvg \cite{mcmahan17afedavg} aggregated model parameters trained locally on devices on a common server with the user data size acting as the aggregation weights. This aggregation method, however, gives more weight to clients with larger data and can cause the global server model to drift towards such clients. Generally, in most works on federated learning, a dataset is split according to a Dirichlet distribution to emulate the imbalance in data distribution or data heterogeneity in real-life scenarios. Besides, data distribution heterogeneity, there can also occur another type of heterogeneity, that is, class heterogeneity, wherein different clients have non-overlapping classes between them \cite{zhuang2022fedema, JIANG2024ufps}. For handling both types of heterogeneity, different frameworks have been proposed over the years \cite{mcmahan17afedavg, zhuang2022fedema, JIANG2024ufps, yang2019FedMLca, Wen2022fedlsurvey}.

%%%%% say application of federated learning is limited in medical image domain %%%%%%
%%% and only gaining traction in recent years %%%%%%%%%

The application of federated learning in the medical image analysis domain is also of utmost importance, as medical data is also privacy-sensitive. The federated learning framework can be understood in simple terms by first considering that there exist multiple institutions like hospitals or clinics, with a separate collection of scans of several patients. However, there may exist patients who have done MR or CT scans at multiple such institutions due to their circumstances. Federated learning-based medical image analysis, especially, medical image segmentation has gained traction over the last few years.

%%%%% introduce the task in this paper that is SSFSS %%%%%

In this work, we aim to implement the one-shot segmentation task in the federated setting. Following CoWPro \cite{manna2024cowpro}, we conduct self-supervised pre-training for the one-shot segmentation task as well to make the pre-training representations aligned with the optimal downstream task representations. The advantage of using the CoWPro method as the baseline is that there is no need for fine-tuning for the downstream task, saving a lot of computational resources. The self-supervised one-shot task involves augmenting a randomly chosen slice from an MR slice in two different ways resulting in the formation of the support and query image and using a random superpixel drawn from the superpixels obtained using the Felzemswalb segmentation algorithm as the foreground mask. In the downstream task, the support and query images are replaced with slices from the held-out support scan and the validation set scans, respectively, while the pseudo-mask is replaced with the ground truth support mask. While there have been works on federated few-shot learning and meta-learning, in general, \cite{song2023fedfsl, tu2024fedfslar, alireza2020perfedavg}, work on federated few-shot learning is scarce. To the best of our knowledge, this work is the first to implement one-shot segmentation is scarce in federated learning scenarios.

Apart from the framework novelty, we also contribute a novel medical imaging dataset. This dataset can be described as a multi-organ gynaecological dataset intended for brachytherapy consisting of magnetic resonance scans from 95 patients who have undergone interstitial as well as intracavitary brachytherapy. While there are existing works on brachytherapy scans on different organs like the cervix, prostate, uterus, and rectum, it is not possible to find a public dataset as large as the one proposed in this work on the web. Additionally, the proposed dataset is public compared to the other works which have used private datasets, and this adds further credibility to our work and is open to public scrutiny.

%%%%%%%% contributions %%%%%%
In short, the contributions of this work can be summarized as follows:

\begin{itemize}
    \item \textbf{Implementation of One-Shot Segmentation in Federated Learning}: To the best of the authors' knowledge, this is the first work to implement one-shot segmentation in federated learning scenarios, addressing a gap in the existing literature on federated few-shot learning. We further use a self-supervised setting which makes the problem more challenging.
    \item \textbf{Introduction of a Novel Medical Imaging Dataset}: The work introduces a new multi-organ gynaecological dataset for brachytherapy, consisting of magnetic resonance scans from 95 patients. This dataset is intended for research in interstitial and intracavitary brachytherapy.
    \item \textbf{Public Availability of the Dataset}: Unlike many existing studies that use private datasets, the proposed dataset is publicly available, enhancing transparency and allowing for public scrutiny.
\end{itemize}

The rest of the work is organized as follows: Section II deals with the literature review, followed by a detailed discussion of the preliminaries in Section III to make it easier for the readers to go through the manuscript. Section IV describes the methodology and the federated learning framework. In Section V, we present the theoretical derivations to prove the convergence guarantees of the proposed frameworks. Section VI details the experimental configurations including the datasets, training and validation details. The experimental results and comparison are presented in Section VII. Finally, we conclude this work in Section VIII.

\section{Literature Review}
\label{sec:litrev}

Before moving on to the methodology of this work, we will review some of the works done in the recent past on federated learning in general and also on the application of federated learning in medical image segmentation.

\subsection{Federated Learning}
\label{subsec:genfedlearning}

Federated learning is currently the benchmark approach for privacy-preserved distributed learning. The prevalent common approach for client model aggregation, named FedAvg was first proposed in \cite{mcmahan17afedavg}. Following this work, several works have been done to improve the aggregation strategy as well as improve representation learning in the clients. FedProx \cite{li2020fedprox} uses a regularization loss to keep the client models from drifting apart from the aggregated server model. FedMA \cite{Wang2020FedMA} proposes constructing the shared global model in a layer-wise manner to deal with the permutation invariance of the neural network parameters. 

FUSL \cite{Zhang2023fusl} applies self-supervised contrastive learning in each client for unsupervised representation learning in the federated learning scenario and also maintains the consistency of representations over the classes. Several other works also applied unsupervised learning frameworks for different applications as in \cite{zhuang2021fedunface1, zhuang2021fedfr, berlo2020towardsfusl}. MOON \cite{li2021moon} uses a model contrastive framework to prevent client drift by using outputs from the server model and the local client model as a positive pair in the contrastive learning framework. FedU \cite{zhuang2021fedu} uses a divergence-aware predictor update as the predictor learns information which is more specific to the clients and helps in striking a balance between the generalizability and unwanted side-effects of updating the local predictor parameters with the global predictor parameters due to large variances. Similar to FedU, FedEMA \cite{zhuang2022fedema} also uses a divergence-aware parameter update for the local client models. FedDecorr \cite{shi2023feddecorr, shi2024feddecorr} states that local client models suffer from dimensional collapse due to the heterogeneity in data distribution and aims to mitigate the same using a decorrelation-based regularization. L-DAWA \cite{rehman2023ldawa} on the other hand, combines the parameters of the parameters of the local client model with the server model based on an angular divergence measure of the parameters. FedACG \cite{kim2024fedacg} uses a framework similar to FedProx, but with a gradient-based momentum term to accelerate the convergence of the aggregated server model.

A federated learning problem can also be posed as a multi-objective optimization problem, where the objective is to find a Pareto stationary solution to the problem. The federated learning approach can be subdivided into two types, weighted approach, such as FedAvg, and Chebyshev approach as in AFL \cite{mohri19afl}, FedMGDA \cite{hu2022fedmgda}, FMGDA \cite{yang2023fmgda}, or FedU2 \cite{liao2024FedU2}, where the approach involves solving a minimax optimization problem to find the Pareto stationary aggregation weights.

% Personalized Federated Learning
% FedBABu, FedBN, perFedAvg, FedLC, Ditto, Floco, pFedMe

\subsection{Federated Learning in Medical Image Segmentation}
\label{subsec:fedlinmedimgseg}

Medical Image Analysis is a privacy-sensitive domain as it involves using the personal data of patients. This makes this domain an important area where federated learning finds applicability with significant importance.

One of the first applications of federated learning in medical image segmentation \cite{Li2019fedbrtumorseg} applied a differentially-private federated learning framework with momentum updated gradient with bias-corrected learning rate for brain tumour segmentation from magnetic resonance scans.

Another work \cite{guo2021multiinst} uses a combination of Fourier-based reconstruction and adversarial training for MR image reconstruction. FedDG \cite{liu2021feddg} implements a frequency domain-based interpolation module for learning representations from multiple source domains by sharing distribution information. FedNorm \cite{Yin2022normten} uses a tensor normalization-based aggregation to account for heterogeneous local client iterations and client pruning to discard slow clients for fast federated training.

LC-Fed \cite{Wang2022LCFed} uses split aggregation of the encoder and the representation head for personalized federated learning for medical image segmentation. FedSM \cite{xu2022fedsm} aims to avoid the client drift issue and close the generalization gap between FL and centralized training for medical segmentation tasks using a trainable model selector. FedMix \cite{wicksana2023fedmix} presents a combination of supervised, weakly-supervised and unsupervised clients using pseudo-label generation and refinement using information from clients with labelled data. FedMENU \cite{Xu2023fedmenu} uses the same number of sub-encoders as the number of organs in the medical image segmentation task with a shared decoder for handling class heterogeneity, which makes the entire framework memory intensive. UPFS \cite{Jiang2024upfs} instead uses pre-trained encoders from other clients to generate pseudo labels for classes except the ones with ground truth in the local client database, combined with an uncertainty-aware aggregation. FedCross \cite{xu2023fedcrossl} aims to eliminate the aggregation process by using a unique mechanism, where the global model is sent to a randomly selected client to train on the local dataset and the updated model is again sent to another randomly selected client in the next communication round.

FedCE \cite{jiang2023fedce} uses a gradient-based contribution of each client to compute aggregation weights. FedIIC \cite{Wu2023fediic}, on the other hand, utilises both intra-client contrastive learning and inter-client contrastive learning through shareable class prototypes, along with difficulty aware logit adjustment to combat the ill-effects of data heterogeneity in the clients.

FedMCIS \cite{Galati2024fedmcis} uses adversarial generative methods combined with an asynchronous domain adaptation step to deal with uneven data distribution due to differences in scanners, modality, and imaged organs. FedA3I \cite{Wu2024feda3i} uses an annotation-quality aware aggregation to deal with heterogeneous noise due to a non-overlapping annotator in each client. Similar to FedMix, in \cite{wu2024fsss}, models trained on labelled clients are used for pseudo-label generation in unlabelled clients, along with the transmission of class prototypes for contrastive pre-training.

In the works above, 

% https://github.com/sadimanna/federated-segmentation
\subsection{Brachytherapy Datasets}

From the literature, we found several applications for brachytherapy data using state-of-the-art machine learning algorithms. However, the dataset proposed in this work is the first public dataset among all the datasets known so far. In Table \ref{tab:data}, we have summarised the descriptions and details of some datasets used in various works for organs-at-risk (OAR) segmentation but not made accessible to the public for use.

\begin{table*}
    \centering
    \caption{Summary of brachytherapy datasets used for segmentation tasks.}
    \begin{tabular}{|c|c|c|c|p{5cm}|c|c|}
    \toprule
        Dataset & Diagnosis & Parts & Imaging Mode & Organs & Samples & Access \\ \midrule
        \cite{Mohammadi2021dlbrachycerv} & Cervical cancer & Cervix & CT & Urinary bladder, rectum, and sigmoid colon & 113 & Private\\ \midrule
        \cite{Zhang2020-ho} & Cervical Cancer & Cervix & CT & small intestine, sigmoid, rectum, and bladder & 91 & Private \\ \midrule
        \cite{Li2022-bc} & Gynaecological Cancer & Cervix &  CT & Bladder, Rectum, HRCTV & 237 & Private \\ \midrule
        \cite{Li2023-qr} & Gynaecological Cancer & Cervix & CT & HRCTV, bladder, rectum &  250 & Private \\ \midrule
        \cite{Yoganathan2022-td} & Cervical Cancer & Cervix & MR &Bladder, Rectum, Sigmoid Colon, small intestine, and HR-CTV & 111 & Private \\ \midrule
        \cite{Wu2025-ni} & Cervical Cancer & Cervix & MR & HR-CTV & 464 & Private \\ \midrule
        \cite{Ni2024-qi} & Cervical Cancer & Cervix & MR & Bladder, Rectum, Sigmoid Colon, Small Bowel & 285 & Private \\ \midrule
        % \cite{Liu2024-jz}  & & & & & & \\ \midrule
        \cite{Xia2025-tc} & Cervical Cancer & Cervix & MR & - & 375 & Private \\ \midrule
        \cite{Zabihollahy2022-us}  & Cervical Cancer & Cervix & MR & Bladder, CTV & 213 & Private\\ \midrule
        \rowcolor{gray!20} MOGaMB & Cervical cancer & Cervical & MR & Urinary Bladder, Rectum, Sigmoid Colon, Femoral Head & 95 & Public\\  \midrule
    \end{tabular}
    \label{tab:data}
\end{table*}

\section{Preliminaries}

% \subsection{Data Collection and Processing}

\subsection{Federated Learning Framework Overview}

Having established the background scenario, let there be a decentralized setting where there are $K$ different sources of data which correspond to different medical institutions in our case, and all store their data locally. The datasets may be independent but non-identically distributed, or in simple terms described as non-IID or heterogeneous in federated learning terminology. 

The goal of federated learning in this scenario is to collaboratively learn a representative model or solve a learning problem by utilizing the data from all the clients. More specifically, let each data source $k \in K$ have a local dataset $\mathcal{D}_k = \{x^i_k, y^i_k\}^{|\mathcal{D}_k|}_{i=1}$, where $x^i_k$ and $y^i_k$ denotes the image and its corresponding segmentation masks.

The federated learning problem can be formulated as finding the set of parameters which minimizes the following objective:

\begin{equation}
    \begin{split}
        \mathcal{L}(\theta, \mathbf{\lambda}) = \sum_{k=1}^K \lambda_k \mathcal{L}_k(\theta, \mathcal{D}_k)
    \end{split}
\end{equation}

where $\mathbf{\lambda} = [\lambda_1, \lambda_2, \hdots, \lambda_K]$ represents the aggregation weights and $\mathcal{L}_k$ is the loss for the $k$-th client. The common aggregation approach FedAvg \cite{mcmahan17afedavg} is to assign the ratio of the number of samples in the $k$-th client to the sum of the number of samples over all the clients as the aggregation weight, that is, $\lambda_k = \frac{|\mathcal{D}_k|}{\sum_{i=1}^K |\mathcal{D}_i|}$.

In our work, we consider a cross-silo federated learning setting, where all the clients are always active. In each communication round $r$, the clients train their local model $w^r_k$ on the local dataset for $E$ local epochs. This trained model is sent to the server, where the server aggregates the client models using the weights $\mathbf{\lambda}$. This aggregated global model $w^r_g$ is then broadcast to the clients. In the next communication round, the client models $w^{r+1}_k$ is initialized with $w^r_g$ for further training.

\subsection{Few-Shot Learning Overview}

In the few-shot learning framework, we need to consider two sets of data, the \textit{Support} set $\mathcal{S}$ and the \textit{Query} set $\mathcal{Q}$. The \textit{Support} set $\mathcal{S}$ consists of the tuple $\{ \mathcal{X}^i_s, \mathcal{Y}^i_s(l) \}^k_{i=1}$, where $\mathcal{X}^i_s$ is the $i$-th sample in the Support set with the segmentation mask $\mathcal{Y}^i_s(l)$ for the class $l$, where $l$ belongs to the set of novel classes available during the testing phase. The primary objective is to learn an approximate function $f$ which takes as input the support set $\mathcal{S}$ and the query image $\mathcal{X}_q$ and predicts the binary mask $\hat{\mathcal{Y}_q}$ of the unseen classes in $\mathcal{X}_q$, denoted by the support mask $\mathcal{Y}_s(l)$. 

The support set $\mathcal{S}$ is a subset of $\mathcal{D}_{train}$. During training, the input to the model is $(\mathcal{S}, \mathcal{X}_q)$. Such a pair is called an \textit{episode}. If during training, the value of $k$ is $1$, that is, we use only a single image in the support set, then the learning is known as one-shot learning, which we adopt in this work. If $k > 1$, it is known as few-shot learning. If the number of classes is $N$, then we call it $N$-way $k$-shot learning.

\subsection{Baseline: CoWPro}

The CoWPro \cite{manna2024cowpro} framework is used as the baseline self-supervised few-shot segmentation framework in our work. The CoWPro framework adopts a prototype-based one-shot segmentation pipeline.

In the self-supervised pre-training step, a random superpixel obtained using the Felzenswalb segmentation algorithm is chosen as the foreground mask. Each input image $\mathcal{X}^i$ is augmented in two different ways to obtain the support image and mask pair $(\mathcal{X}^i_s, \mathcal{Y}^i_s)$ and the query image $\mathcal{X}^i_q$ and the ground truth mask $\mathcal{Y}^i_q$. After extracting the support and query image features $f(\mathcal{X}^i_s)$ and $f(\mathcal{X}^i_q)$, the foreground prototypes $\mathcal{P}^f_q$ and background prototypes $\mathcal{P}^b_q$ are obtained by elementwise multiplication of the downsampled foreground and background mask, that is, $\mathcal{P}^f_q = f(\mathcal{X}^i_s) \odot \mathcal{Y}^i_s$ and $\mathcal{P}^b_q = f(\mathcal{X}^i_s) \odot \bar{\mathcal{Y}^i_s}$, where $\bar{\mathcal{Y}^i_s}$ denotes the complement of the binary mask $\mathcal{Y}^i_s$. Using the prototypes $\mathcal{P}^f_q$ and $\mathcal{P}^b_q$, a score map is generated for each query pixel $f(\mathcal{X}^i_q)$ where a score is assigned to each of the prototypes in $\mathcal{P}^f_q$ and $\mathcal{P}^b_q$. Next, two prototypes, one for the foreground and background, are generated for each query pixel. Finally, in the last step, the cosine similarity of both the weight-based aggregated foreground and background prototype with each query pixel is calculated, which serves as the final prediction score to decide whether the pixel belongs to the foreground or background.

The advantage of the CoWPro framework is that it does not require fine-tuning the true ground truth data in the downstream task of one-shot segmentation. The pipeline remains the same, while only the support and query inputs are changed. This framework also does not use a separate decoder for predicting the segmentation masks, which makes this framework computationally efficient, and better suited for a federated learning scenario.

\section{Methodology and Study Overview}

In this section, we will discuss the entire framework used in this work. First, we discuss the problem statement of the framework adopted in this work, followed by the overview of the framework and the reinforcement adopted to improve the baseline federated framework.

\subsection{FedCowPro Problem Statement}
\label{ssec:probst}

The problem statement for our proposed FedCoWPro can be described as a federated version of the self-supervised few-shot segmentation framework CoWPro. The objective is to devise a federated learning framework which is capable of assimilating information from different clients or data sources and broadcast the aggregated information among the clients such that the collective performance of the clients improves. 

However, in our work, the problem statement is more challenging than conventional federated segmentation frameworks \cite{} as we attempt to implement a very strict \textit{data annotation scarcity} scenario, assuming that only one scan among the available scans in each client is annotated, leading directly to the motivation for the \textit{one-shot segmentation} task undertaken in this work. To deal with such data annotation scarcity, it is imperative to utilise information from other data sources, such that the model can learn better representations of the data, and help in more robust downstream performance. 

The motivation for adopting a self-supervised learning framework can be traced to the recent work \cite{wang2023does} which evidences that decentralized self-supervised learning helps in learning robust representations and is also effective in the face of data heterogeneity.

\subsection{Overview of the FedCoWPro Framework}

%%%% small plot comparing vanilla version with federated version %%%%%%%

The FedCoWPro framework consists of two parts: (1) Clients, (2) Server. Inside each client, have a local CoWPro model and a local dataset which can be of MR or CT imaging technique.

Following the CoWPro \cite{manna2024cowpro} framework, we generate pseudo-segmentation masks from the local datasets and use it to train the local encoder in each client. After $E$ local epochs of pre-training, the client's local models are sent to the server for aggregation. The server aggregates the weights and broadcasts them back to each client. 

We also evaluate the aggregated models on the server using an unseen CT or MR dataset. This helps us to evaluate the quality of self-supervised representation learning in our framework and the robustness of decentralized learning. For the aggregation, we use the simple FedAvg \cite{mcmahan17afedavg} framework.

The aggregation weights are computed based on the number of slices used for each client. However, due to the large number of slices in the CT datasets, we also tried capping the number of iterations in each local epoch to 1000 to prevent bias towards the clients with CT datasets, and the aggregation weights were computed accordingly. An undesired bias towards the CT clients may hamper the convergence of clients with MR datasets and prevent the overall performance of the clients and the server on the downstream segmentation task.

\begin{figure*}
    \centering
    \includegraphics[width=0.95\linewidth]{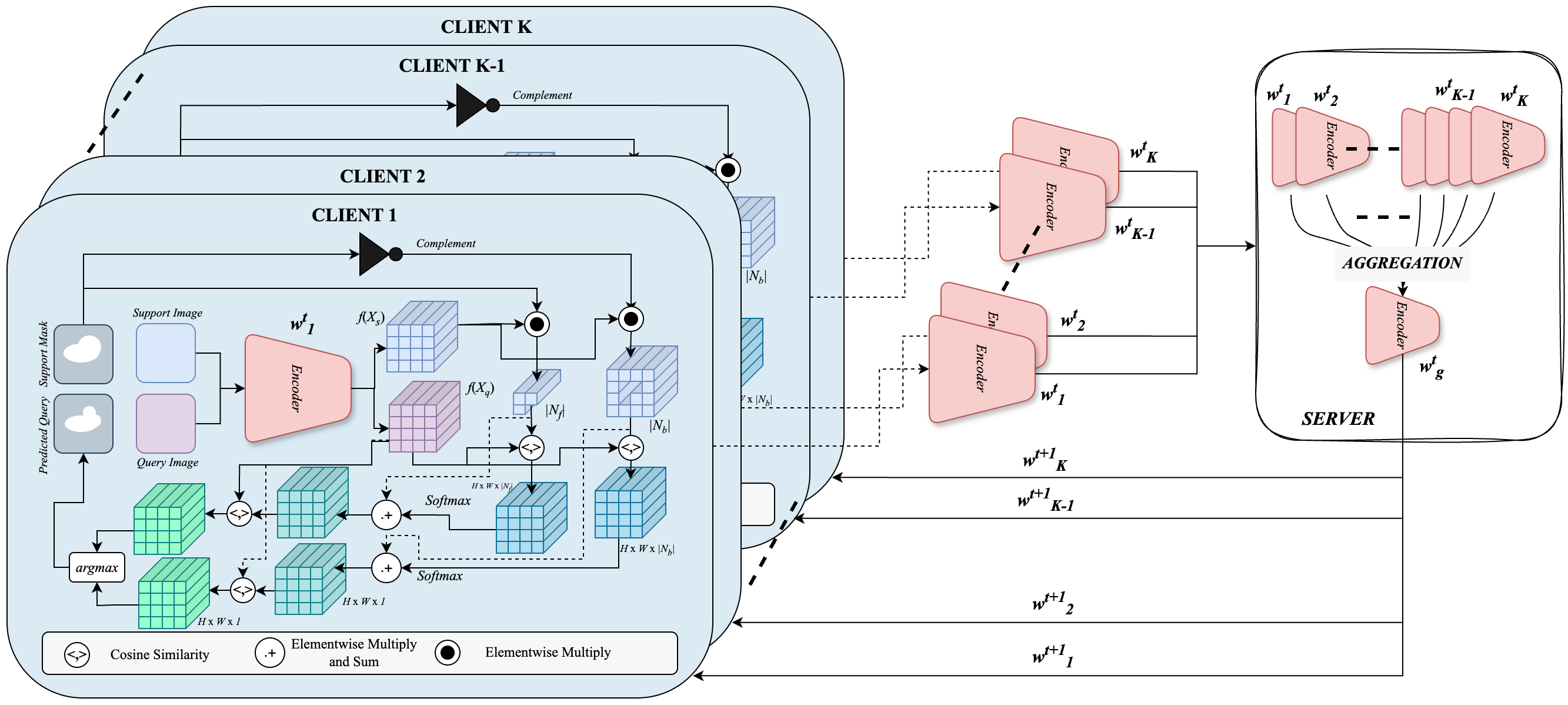}
    \caption{An illustrative diagram showing the FedCoWPro framework.}
    \label{fig:fedcowpro}
\end{figure*}

\subsection{Reinforcing with DICE Loss}
\label{sssec:diceloss}

The baseline CoWPro framework optimizes a weighted Cross-Entropy loss and a cyclic consistency loss to optimise the model parameters. To improve the performance of the baseline CoWPro framework, we reinforce the CoWPro loss, by adding two losses, namely, Spatial Dice loss and Edge Dice loss, to improve segmentation mask prediction. In the following paragraphs, we dive into the details of the loss variants. Interestingly, the formulation of the loss function for both variants is the same, but the entities on which it is applied vary.

\paragraph{Spatial Dice Loss}

The spatial dice loss \cite{sudre2017gendiceloss} takes the entire ground truth and predicted mask into consideration when calculating the dice loss. Let $p_{true}$ be the ground truth mask and $p_{pred}$ be the predicted mask. Then the spatial dice loss is defined as follows,

\begin{equation}
    \mathcal{L}_{ds}(p_{true},p_{pred}) = 1 - \frac{2 \times p_{true} \times p_{pred}}{\sum p_{true} + \sum p_{pred} + \epsilon}
    \label{eqn:ds}
\end{equation}

where $p_{pred}$ and $p_{true}$ are the predicted and ground truth query mask, respectively.

\paragraph{Edge Dice Loss}

\begin{figure}
    \centering
    \includegraphics[width=0.9\linewidth]{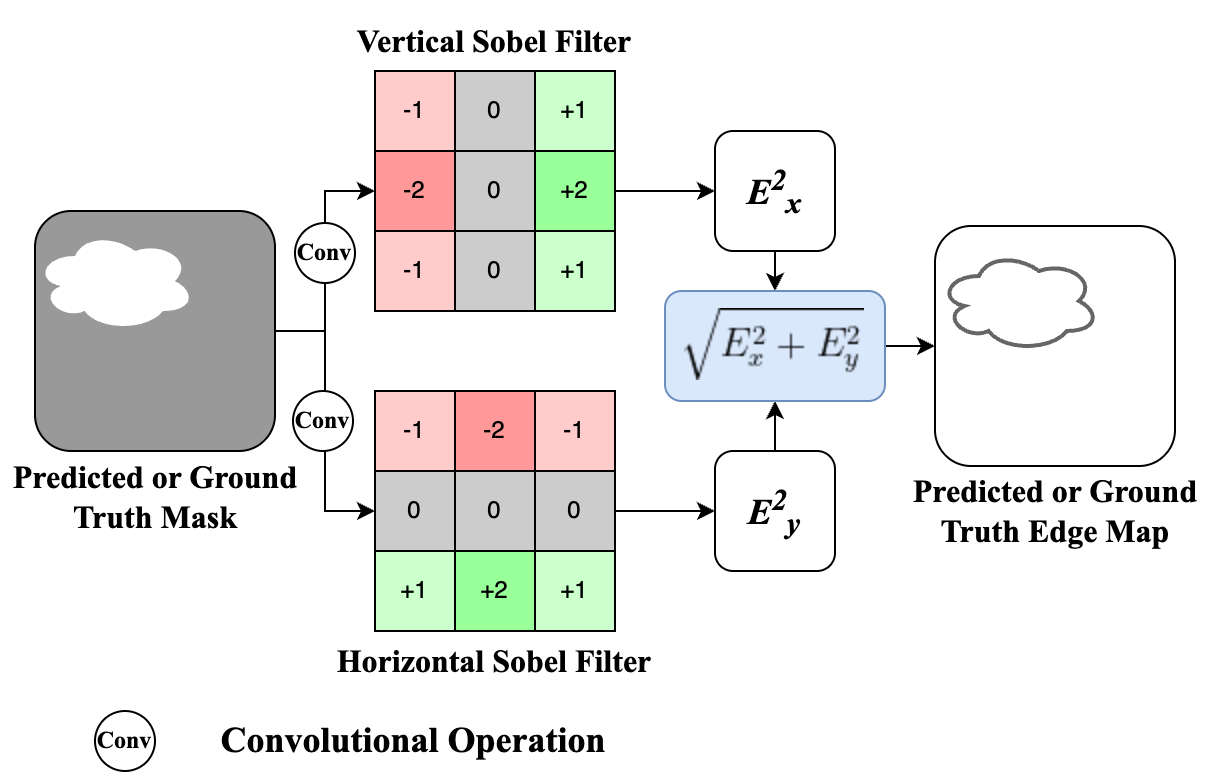}
    \caption{Pipeline for computing edge maps of the ground truth and predicted query mask.}
    \label{fig:edgedice}
\end{figure}

In literature, the edge dice loss \cite{deng2018crisp} has been particularly used to predict edge maps. However, the formulation in \cite{deng2018crisp} may run into numerical instability occasionally due to the nature of its formulation. Hence, we use a different formulation of edge dice loss more aligned to the conventional dice loss to enforce better prediction of the query mask. The formulation of the edge dice loss used in our work is the same as in Eqn. \ref{eqn:ds}, but we present it below for the ease of the readers. However, the computation of the edge dice loss involves convolving the ground truth and predicted mask with horizontal and vertical sobel filters and then aggregating them to compute the respective edge maps as shown in Fig. \ref{fig:edgedice}. The edge dice loss is then computed on the predicted and ground truth edge maps as shown in Eqn. \ref{eqn:es}.

\begin{equation}
    \mathcal{L}_{es}(e_{true}, e_{pred}) = 1 - \frac{2 \times e_{true} \times e_{pred}}{\sum e_{true} + \sum e_{pred} + \epsilon}
    \label{eqn:es}
\end{equation}

where $e_{true}$ and $e_{pred}$ are the edge maps of the ground truth and the predicted query masks.

\paragraph{Final Loss}

The expression of the combined loss can be defined as follows,

\begin{equation}
\begin{split}
    \mathcal{L}_{total}((\mathcal{X}^i_s, \mathcal{Y}^i_s), \mathcal{X}^i_q, \hat{\mathcal{Y}}^i_q) &= \mathcal{L}_{cowpro}((\mathcal{X}^i_s, \mathcal{Y}^i_s), \mathcal{X}^i_q, \hat{\mathcal{Y}}^i_q) \\
    &+\mathcal{L}_{ds}(\mathcal{Y}^i_s, \hat{\mathcal{Y}}^i_q) + \mathcal{L}_{es}(\mathcal{E}(\mathcal{Y}^i_s), \mathcal{E}(\hat{\mathcal{Y}}^i_q))
\end{split}
\end{equation}

where $\mathcal{E}$ denotes the sobel edge operation.
%%%% give plots of improvement here %%%%

%\section{Convergence Guarantees}

\section{Experimental Details}

In this section, we describe the implementation details and the experimental configuration for obtaining the results provided in this manuscript.

\subsection{Datasets}

\subsubsection{\textbf{M}ulti-\textbf{O}rgan \textbf{G}yn\textbf{a}ecological \textbf{M}R \textbf{B}rachytherapy (MOGaMB) Dataset}

In Sec. \ref{sec:intro}, we mentioned that we introduce a novel magnetic resonance dataset obtained from patients undergoing gynaecological brachytherapy as an important contribution to this manuscript. In the following paragraphs, we present the data collection, image acquisition, data pre-processing strategies and other details.

\paragraph{Data selection} In general, for locally advanced cervical cancer, the International Federation of Gynecology and Obstetrics (FIGO) guidelines recommend including brachytherapy as a standard treatment protocol for stages IB2-IVA. We consider 95 patients who have undergone interstitial as well as intracavitary brachytherapy for our study. All procedures were performed under spinal sedation. To maintain consistent bladder volume, which affects dosimetry, a Foley catheter was inserted into the patients’ bladders through the urethra.

\paragraph{Image acquisition} All 95 patients with applicator insertions underwent pre-treatment scans using a clinical MR scanner (GE Signa Explorer 1.5T) equipped with a phase array body coil. A high-resolution T2-weighted (T2W) scan was chosen for its superior soft-tissue contrast and relatively fast imaging acquisition. The image acquisition time ranged from 160 to 180 seconds, with 3-dimensional scanning. The repetition time was 2600 ms, and the echo time was 155 ms. The imaging method included 30 to 50 slices per scan, each of size 512 × 512 pixels, with a pixel resolution of 0.5 mm × 0.5 mm.

\paragraph{Data preprocessing} For each dataset, the urinary bladder, rectum, sigmoid colon
and femoral head were manually contoured by the expert Radiation Oncologist in MR
image interpretation, using an open-source library ITK-SNAP \cite{itksnap}. It is designed to assist
with the manual segmentation of medical images, such as MRI scans. The interface
is user-friendly, enabling visualization and manipulation of medical images, and allows
users to define or trace regions of interest (ROI). It is frequently utilized in the creation
of medical imaging software and algorithms.

\paragraph{Data Deidentification and File Type Conversion Data} Deidentification is crucial
for organizations to safeguard patient demographic information in data that is collected,
utilized, archived, and shared with other entities. This process involves removal of
explicit identifiers about the health conditions and identity of the patients present
in the data. Deidentification measures are essential to safeguard individual privacy.
Once a dataset is deidentified and no longer contains personal information, its use or
disclosure cannot compromise individual privacy.

\paragraph{Class Label Information} In this dataset, we annotated four organs, namely, Urinary bladder, Rectum, Sigmoid Colon, and  Femoral head. These organs were carefully annotated by multiple expert medical personnel from renowned medical institutions with several years of experience. 

%%%% insert a few samples here %%%%%

\subsubsection{\textbf{C}ombined \textbf{H}ealthy \textbf{A}bdominal \textbf{O}rgan \textbf{S}egmentation (CHAOS) Dataset \cite{CHAOSdata2019}} 

The CHAOS dataset \cite{CHAOSdata2019} includes 120 DICOM data sets from two different MRI sequences, namely, T1-DUAL in phase (40 scans), out phase (40 scans) and T2-SPIR (40 scans). This dataset contains annotations for four organs in the abdomen, liver, left and right kidneys, and spleen. The scans in this dataset are acquired by a 1.5T Philips MRI, which produces 12-bit DICOM images having a resolution of 256 x 256. As per the dataset information from the CHAOS challenge \cite{CHAOS2021}, the ISDs vary between 5.5-9 mm (average 7.84 mm), x-y spacing is between 1.36 - 1.89 mm (average 1.61 mm) and the number of slices is between 26 and 50 (average 36). In total, 1594 slices (532 slices per sequence) will be provided for training and 1537 slices will be used for the tests. Since T1-DUAL is a T1-weighted sequence, it is very effective to identify blood and tissues that are rich in protein. SPIR (Spectral Pre-Saturation Inversion Recovery) stands for a hybrid imaging sequence and uses a T2-weighted contrast mechanism.

\subsubsection{Multi-Atlas Labeling \textbf{B}eyond \textbf{T}he \textbf{C}ranial \textbf{V}ault (BTCV) Dataset \cite{landmanbvc}}

The original BTCV dataset consists of 50 abdomen CT scans randomly selected from a combination of an ongoing colorectal cancer chemotherapy trial, and a retrospective ventral hernia study. The 50 scans were captured during portal venous contrast phase with variable volume sizes ($512 \times 512 \times 85 - 512 \times 512 \times 198$) and field of views (approx. $280 \times 280 \times 280 mm^3 - 500 \times 500 \times 650 mm^3$). The in-plane resolution varies from $0.54 \times 0.54 mm^2$ to $0.98 \times 0.98 mm^2$, while the slice thickness ranges from 2.5 mm to 5.0 mm. Thirteen abdominal organs were manually labelled by two experienced undergraduate students, and verified by a radiologist on a volumetric basis using the MIPAV software, including the spleen, right kidney, left kidney, gallbladder, oesophagus, liver, stomach, aorta, inferior vena cava, portal vein and splenic vein, pancreas, right adrenal gland, and left adrenal gland. However, for the segmentation task in this work, we only use four labels, namely, liver, left and right kidneys, and spleen.

\subsubsection{\textbf{F}ast and \textbf{L}ow-resource semi-supervised \textbf{A}bdominal o\textbf{R}gan s\textbf{E}gmentation (FLARE 2022 or FLARE22) Dataset \cite{FLARE22-LDH2024}}

The FLARE22 dataset is also a Computed Tomography dataset. This dataset contains a small number of labelled cases (50) called the tuning set and a large number of unlabeled cases (2000) in the training set, 50 visible cases for validation, and 200 hidden cases for testing. The segmentation targets include 13 organs: liver, spleen, pancreas, right kidney, left kidney, stomach, gallbladder, oesophagus, aorta, inferior vena cava, right adrenal gland, left adrenal gland, and duodenum. For this work, we only use the tuning set. Furthermore, we divide the 50 scans in the tuning set into two parts, consisting of 30 and 20 scans and allow us to increase the number of clients in our experiments.

% \subsubsection{\textbf{F}ast, \textbf{L}ow-resource, and \textbf{A}ccurate o\textbf{R}gan and Pan-cancer s\textbf{E}gmentation (FLARE 2023 or FLARE23) Dataset \cite{ma2024flare23, flare23}}

% The FLARE23 dataset is even larger than FLARE22. The training set includes 4000 3D CT scans from 30+ medical centers. 2200 cases have partial labels and 1800 cases are unlabeled. However, due to computational constraints, we only select the first 75 scans from the training set, and divide the 75 scans into 3 parts constituting 3 clients in our proposed federated learning framework. The 3 clients were distributed 30, 20, and 25 scans among them, with no overlap between them. The FLARE23 dataset also provides the labels for the same 13 organs as in the FLARE22 dataset.

\subsection{Server-Client Dataset Distribution}

To prove the plausibility of our proposed framework, we used 5 datasets, which were divided into nine subdivisions, named MOGaMB, CHAOST2, CHAOST1\_In, CHAOST1\_Out, BTCV, FLARE22\_1, FLARE22\_2. %, FLARE23\_1, FLARE23\_2, FLARE23\_3. 
Among these seven parts, the first four are magnetic resonance (MR) datasets, and the rest are computed tomography (CT) datasets. This enables us to test our proposed framework on a combination of datasets with varying data distribution satisfying the condition of non-IID data in a real-life federated learning scenario.

To test our proposed federated learning framework, we use a held-out validation set in each client. 

We conduct our experiments on abdominal and gynaecological imaging data separately. For the first experimental setting, we first divide the MOGaMB dataset into five parts, where each part will constitute a client. Each of the five clients contains 20, 20, 15, 15 and 25 samples respectively. In the second setting, we consider only the datasets CHAOST2, CHAOST1\_In, CHAOST1\_Out, BTCV, FLARE22\_1, and FLARE22\_2 as the clients.

%%%%%%%%%%%%%%%%%%%%%%%%%%%%%%%%%%%%%%%%%%%%%%%%%%%%%%%%%%%%%%%%%%%
\begin{table*}[!ht]

    \centering
    \caption{Comparison of the results of the CoWPro framework applied in the Federated learning scenario with the FedAvg aggregation strategy on the proposed dataset MOGaMB.}
    \begin{tabular}{|c|c|c|c|c|c|c|c|c|c|}
    \toprule
        \multirow{2}{*}{Method} & \multirow{2}{*}{Client}&\multicolumn{4}{c|}{Losses} & \multicolumn{4}{c|}{DICE score} \\ \cline{3-10}
         & & Cross-Entropy & Regularization & Spatial Dice & Edge Dice & U.Bladder & Rectum & Sig. Colon & Fem. Head \\ \midrule 
         FedAvg+CoWPro & \multirow{2}{*}{1} & $\checkmark$ & $\checkmark$ & $\times$ & $\times$ & \textbf{45.92} & \textbf{59.41} & \textbf{37.94} & \textbf{41.54} \\
         FedAvg+CoWPro & & $\checkmark$ & $\checkmark$ & $\checkmark$ & $\checkmark$ & 44.59 & 53.11 & 34.96 & 39.65 \\ \midrule
         %%%%%%%%%%%%%%%%%%%%%%%%%%%%%%%%%%%%
         FedAvg+CoWPro & \multirow{2}{*}{2} & $\checkmark$ & $\checkmark$ & $\times$ & $\times$ & 51.04 & \textbf{64.31} & \textbf{23.24} & \textbf{60.67} \\
         FedAvg+CoWPro & & $\checkmark$ & $\checkmark$ & $\checkmark$ & $\checkmark$ & \textbf{51.09} & 57.94 & 23.04 & 59.69 \\ \midrule
         %%%%%%%%%%%%%%%%%%%%%%%%%%%%%%%%%%%%
         FedAvg+CoWPro & \multirow{2}{*}{3} & $\checkmark$ & $\checkmark$ & $\times$ & $\times$ & \textbf{56.05} & \textbf{47.71} & \textbf{42.56} & \textbf{72.83} \\
         FedAvg+CoWPro & & $\checkmark$ & $\checkmark$ & $\checkmark$ & $\checkmark$ & 54.94 & 46.86 & 40.30 & 70.79 \\ \midrule
         %%%%%%%%%%%%%%%%%%%%%%%%%%%%%%%%%%%%
         FedAvg+CoWPro & \multirow{2}{*}{4} & $\checkmark$ & $\checkmark$ & $\times$ & $\times$ & 36.11 & 48.99 & 36.27 & \textbf{58.27} \\
         FedAvg+CoWPro & & $\checkmark$ & $\checkmark$ & $\checkmark$ & $\checkmark$ & \textbf{36.13} & \textbf{50.81} & \textbf{36.42} & 55.67 \\ \midrule
         %%%%%%%%%%%%%%%%%%%%%%%%%%%%%%%%%%%%
         FedAvg+CoWPro & \multirow{2}{*}{5} & $\checkmark$ & $\checkmark$ & $\times$ & $\times$ & 51.67 & \textbf{39.76} & \textbf{51.55} & 69.74 \\
         FedAvg+CoWPro & & $\checkmark$ & $\checkmark$ & $\checkmark$ & $\checkmark$ & \textbf{52.44} & 34.79 & 38.74 & \textbf{72.02}\\ %\midrule
         %Supervised & \multirow{2}{*}{Centralized}  & $\checkmark$ & $\times$ & $\times$ & $\times$ & 0.9231 &0.8773 &0.7625 & 0.8935 \\ \cmidrule{3-10} 
         %(not one shot) &  & $\times$ & $\times$ & $\checkmark$ & $\times$ & 0.9003&0.8435&0.7441&0.8781 \\ 
         \bottomrule
    \end{tabular}
    \label{tab:gynaemrtab}
\end{table*}

\begin{table*}[!ht]
    \centering
    \caption{Comparison of the results of the CoWPro framework applied in the Federated learning scenario with the FedAvg aggregation strategy on the datasets BTCV, FLARE22\_1, FLARE22\_2, CHAOST1\_In, CHAOST1\_Out, CHAOST2, each acting as a separate client.}
    \begin{tabular}{|c|c|c|c|c|c|c|c|c|c|}
    \toprule
        \multirow{2}{*}{Method} & \multirow{2}{*}{Client}&\multicolumn{4}{c|}{Losses} & \multicolumn{4}{c|}{DICE score} \\ \cline{3-10}
         & & Cross-Entropy & Regularization & Spatial Dice & Edge Dice & Liver & Right K. & Left K. & Spleen \\ \midrule 
         FedAvg+CoWPro & \multirow{2}{*}{BTCV} & $\checkmark$ & $\checkmark$ & $\times$ & $\times$ & 67.56 & 67.1& 82.75 & 68.39\\
         FedAvg+CoWPro &  & $\checkmark$ & $\checkmark$ & $\checkmark$ & $\checkmark$ & 66.71 & 58.56 &  79.81 & 70.32\\ \midrule
         FedAvg+CoWPro & \multirow{2}{*}{FLARE22\_1} & $\checkmark$ & $\checkmark$ & $\times$ & $\times$ & 78.47 & 88.40 & 87.66 & 85.11 \\
         FedAvg+CoWPro &  & $\checkmark$ & $\checkmark$ & $\checkmark$ & $\checkmark$ & 78.23 & 89.89 & 85.02 & 81.23\\\midrule
         FedAvg+CoWPro & \multirow{2}{*}{FLARE22\_2} & $\checkmark$ & $\checkmark$ & $\times$ & $\times$ & 87.21 & 81.12 & 86.37 & 85.05\\
         FedAvg+CoWPro &  & $\checkmark$ & $\checkmark$ & $\checkmark$ & $\checkmark$ & 87.64 & 80.21 & 85.99 & 81.10 \\\midrule
         FedAvg+CoWPro & \multirow{2}{*}{CHAOST2} & $\checkmark$ & $\checkmark$ & $\times$ & $\times$ & 71.20 & 84.06 & 84.12 & 65.97 \\
         FedAvg+CoWPro &  & $\checkmark$ & $\checkmark$ & $\checkmark$ & $\checkmark$ & 72.99 & 81.67 & 85.18 & 66.18\\\midrule
         FedAvg+CoWPro & \multirow{2}{*}{CHAOST1\_Out} & $\checkmark$ & $\checkmark$ & $\times$ & $\times$ & 76.48 & 57.01 & 76.55 & 49.30\\
         FedAvg+CoWPro &  & $\checkmark$ & $\checkmark$ & $\checkmark$ & $\checkmark$ & 73.54 & 59.95 & 75.44& 49.85\\\midrule
         FedAvg+CoWPro & \multirow{2}{*}{CHAOST1\_In} & $\checkmark$ & $\checkmark$ & $\times$ & $\times$ & 74.36 & 63.11 & 75.94 & 52.80\\
         FedAvg+CoWPro &  & $\checkmark$ & $\checkmark$ & $\checkmark$ & $\checkmark$ & 73.56& 52.21& 76.84& 51.34\\\midrule
    \end{tabular}
    \label{tab:abdmrtab}
\end{table*}

%%%%%%%%%%%%%%%%%%%%%%%%%%%%%%%%%%%%%%%%%%%%%%%%%%%%%%%%%%%%%%%

\subsection{Implementation Details}

The proposed framework was implemented purely using PyTorch \cite{Ansel_PyTorch_2_Faster_2024} and its associated libraries like torchvision \cite{TorchVision}. Following CoWPro \cite{manna2024cowpro} and ALPNet \cite{ouyang202sslalpnet}, we used the MS-COCO \cite{lin2014mscoco} pretrained DeepLabv3 ResNet101 backbone \cite{chen2017deeplabv3aspp}, but without the ASPP module \cite{chen2018deeplab}, as it was observed that the segmentation performance dropped when using it \cite{ouyang202sslalpnet}.

Each client was trained for 1 local epoch in each of 100 communication rounds between the server and the clients. 
In each local epoch, the local client model was trained with a batch size of 1 and the model parameters were trained by optimizing a combination of pixel prediction cross-entropy loss, a cyclic consistency loss, spatial dice loss and edge dice loss as described in Sec. \ref{sssec:diceloss} using a SGD optimizer with a learning rate of 1e-3. The learning rate was decayed every epoch by a multiple of 0.96.

% For aggregating the client models, we used the vanilla FedAvg \cite{mcmahan17afedavg} method. Using the ratio of the total number of slices in each client dataset as the aggregation weight for each client, the server model is obtained.

Throughout this manuscript, we use the one-shot segmentation framework, that is, only one pair of support images and mask was used for predicting the query mask during both training and inference.

\subsection{Training and Validation Details}

Before the training commences, the client dataset is divided into a training and validation set in the ratio of 4:1. We take one scan from the training set as the support scan and do not use it in training. Thus, if the number of scans in a dataset is $5N$, then $4N-1$ scans are used for training and $N$ scans are used for validation, while a single scan is used for the one-shot evaluation. 

During validation, each scan, including the support scan, is divided into 3 equal parts. The middle slice from each part is selected as the support image, and the corresponding mask as the support mask. This middle slice from each part is used for one-shot segmentation of a particular label for all the slices in the corresponding part of the query scan. More validation details can be found in \cite{manna2024cowpro}.

\section{Experimental Results}

% \subsection{Comparison with Centralized SSFSS Frameworks}

\subsection{Comparison with Federated Segmentation Frameworks}

In this section, we present the comparison of the proposed federated learning framework on the proposed dataset MOGaMB, a gynaecological brachytherapy dataset split into five parts, each part acting as a separate client. We also conduct experiments on the abdominal datasets, BTCV, FLARE22\_1, FLARE22\_2, CHAOST2, CHAOST1\_In, and CHAOST1\_Out, each acting as separate clients. 

The experimental results are Tables \ref {tab:gynaemrtab} and \ref{tab:abdmrtab}. In \ref {tab:gynaemrtab}, the proposed algorithm is applied on MOGaMB for different clients. The performance of the proposed algorithm is evaluated using six other abdominal datasets as separate clients. 

% \subsection{Qualitative Segmentation Analysis}

% \section{Ablation Study}

% \subsection{Ablation on Dice loss co-efficients $\lambda_{Ds}$ and $\lambda_{De}$}

% \subsection{Ablation on averaging window dimensions}

% \subsection{Ablation on backbone capacity}

% \subsection{Ablation on initialization weights: MS-COCO vs. ImageNet}

% \subsection{Ablation on ASPP module}

% \subsection{Ablation on Federated learning framework}

\section{Conclusion}
Decentralized federated learning enables us to train deep models without compromising privacy. In this work, we developed a federated SSL model in a few-shot settings for medical image segmentation tasks. We considered that the clients had data from different image modalities to make the problem even more realistic. We observed that our model significantly improved the baseline even on an unseen held-out validation set. Being one of the first works in few-shot SSL in distributed settings, we believe this research will open avenues for further research. 
% \section{References}

\bibliographystyle{ieeetr}
\bibliography{main}

\end{document}